\documentclass[runningheads]{llncs}
\usepackage[T1]{fontenc}
\usepackage{graphicx}

\usepackage{multirow}
\usepackage{booktabs}
\begin{document}
\title{Ensemble of ConvNeXt V2 and MaxViT for Long-Tailed CXR Classification with View-Based Aggregation}
\titlerunning{Ensemble for CXR Classification with View-Based Aggregation}
\author{Yosuke Yamagishi\inst{1} \and Shouhei Hanaoka\inst{2}}
\authorrunning{Y. Yamagishi and S. Hanaoka}
\institute{The University of Tokyo, Tokyo 113-8655, Japan \\
\email{yamagishi-yosuke0115@g.ecc.u-tokyo.ac.jp} 
\and The University of Tokyo Hospital, Tokyo, Japan \\
\email{hanaoka-tky@g.ecc.u-tokyo.ac.jp} 
}

\maketitle             % typeset the header of the contribution
\begin{abstract}
In this work, we present our solution for the MICCAI 2024 CXR-LT challenge, achieving 4th place in Subtask 2 and 5th in Subtask 1. We leveraged an ensemble of ConvNeXt V2 and MaxViT models, pretrained on an external chest X-ray dataset, to address the long-tailed distribution of chest findings. The proposed method combines state-of-the-art image classification techniques, asymmetric loss for handling class imbalance, and view-based prediction aggregation to enhance classification performance. Through experiments, we demonstrate the advantages of our approach in improving both detection accuracy and the handling of the long-tailed distribution in CXR findings. The code is available at \url{https://github.com/yamagishi0824/cxrlt24-multiview-pp}.


\end{abstract}
\section{Introduction}
The classification of chest X-rays (CXR) is a complex task due to the variety of possible findings, many of which follow a long-tailed distribution ~\cite{holste2024towards}. Some conditions are rare, yet their accurate detection is crucial in clinical practice. Furthermore, CXR studies often consist of multiple views (frontal and lateral), each providing unique information that must be integrated for a reliable diagnosis. This paper introduces an ensemble method that tackles these challenges by employing ConvNeXt V2 ~\cite{liu2022convnet} and MaxViT ~\cite{tu2022maxvit}, state-of-the-art models known for their powerful feature extraction capabilities. We further address the class imbalance problem with an asymmetric loss function ~\cite{ridnik2021asymmetric} and enhance overall performance using view-based prediction aggregation.

\section{Methods}

\subsection{Dataset and Tasks}

The CXR-LT 2024 challenge expands upon the MIMIC-CXR dataset ~\cite{johnson2019mimiccxrjpglargepubliclyavailable}, featuring 377,110 CXR images annotated with 45 binary disease labels. The training set consists of over 250,000 CXR images with 40 binary disease labels. We participated in two tasks:

\subsubsection{Task 1: Long-tailed Classification on Large, Noisy Test Set}
This task evaluates models on a large-scale, automatically labeled test set of over 75,000 CXR images, annotated with the same 40 binary disease labels as the training set. It assesses performance in a setting that mimics real-world data distribution and labeling noise.

\subsubsection{Task 2: Long-tailed Classification on Small, Manually Annotated Test Set}
Models are evaluated on a "gold standard" subset of 409 manually annotated CXR images, covering 26 out of the 40 disease labels. This provides a more reliable assessment of model performance, particularly for rare conditions.

Both tasks use macro-averaged mean Average Precision (mAP) as the evaluation metric, providing a balanced assessment across all disease categories, including rare conditions.

\subsection{Model Architecture}
ConvNeXt V2 and MaxViT were selected for their complementary strengths in feature extraction. ConvNeXt V2, with its convolution-based architecture, excels at capturing local features, making it effective for detecting fine patterns in chest X-rays. In contrast, MaxViT employs a hybrid design, combining convolutional and transformer modules, allowing it to learn both local and global features. This dual capability is particularly advantageous for capturing the complex structures often present in chest radiographs, such as lung opacities or nodules.

\subsection{Domain-Specific Pretraining}
To adapt the models to the medical imaging domain, we pretrained them on the NIH Chest X-ray dataset, which consists of 112,120 images labeled for 14 common thoracic conditions. This dataset was split into a 19:1 ratio for training and validation. Pretraining on NIH CXR data allows the models to learn domain-specific features, improving their ability to recognize subtle patterns indicative of various chest pathologies.

\subsection{Addressing Class Imbalance}
The CXR-LT dataset exhibits a long-tailed distribution, where common conditions like cardiomegaly are overrepresented compared to rarer findings like pneumothorax. To address this, we employed an asymmetric loss function that assigns higher weights to rare classes, thereby mitigating the bias toward frequent classes. This loss function also employs asymmetric focusing, emphasizing difficult-to-classify cases by adjusting the focus differently for positive and negative samples.

\subsubsection{Implementation Details}
We implemented the asymmetric loss function based on by the first-place solution in the CXR-LT 2023 competition ~\cite{kim2023chexfusion}, with modifications to suit our specific dataset.

The effectiveness of this approach was inspired by its success in last year's competition, where it significantly improved the detection of rare findings.

\subsection{Multi-View Aggregation}
Chest X-ray studies often include images from multiple views. We adopted a view-based prediction aggregation method, where predictions from frontal and lateral views are averaged before applying a weighted aggregation to obtain the final study-level prediction. This view-based prediction aggregation approach was validated by the first-place solution  in the CXR-LT 2023 ~\cite{kim2023chexfusion}. As the 7th place team, we employed a similar technique, recognizing its effectiveness in improving overall prediction accuracy ~\cite{yamagishi2023effect}.

\subsubsection{Aggregation Strategy}
\begin{figure}[ht]
    \centering
    \includegraphics[width=0.8\textwidth]{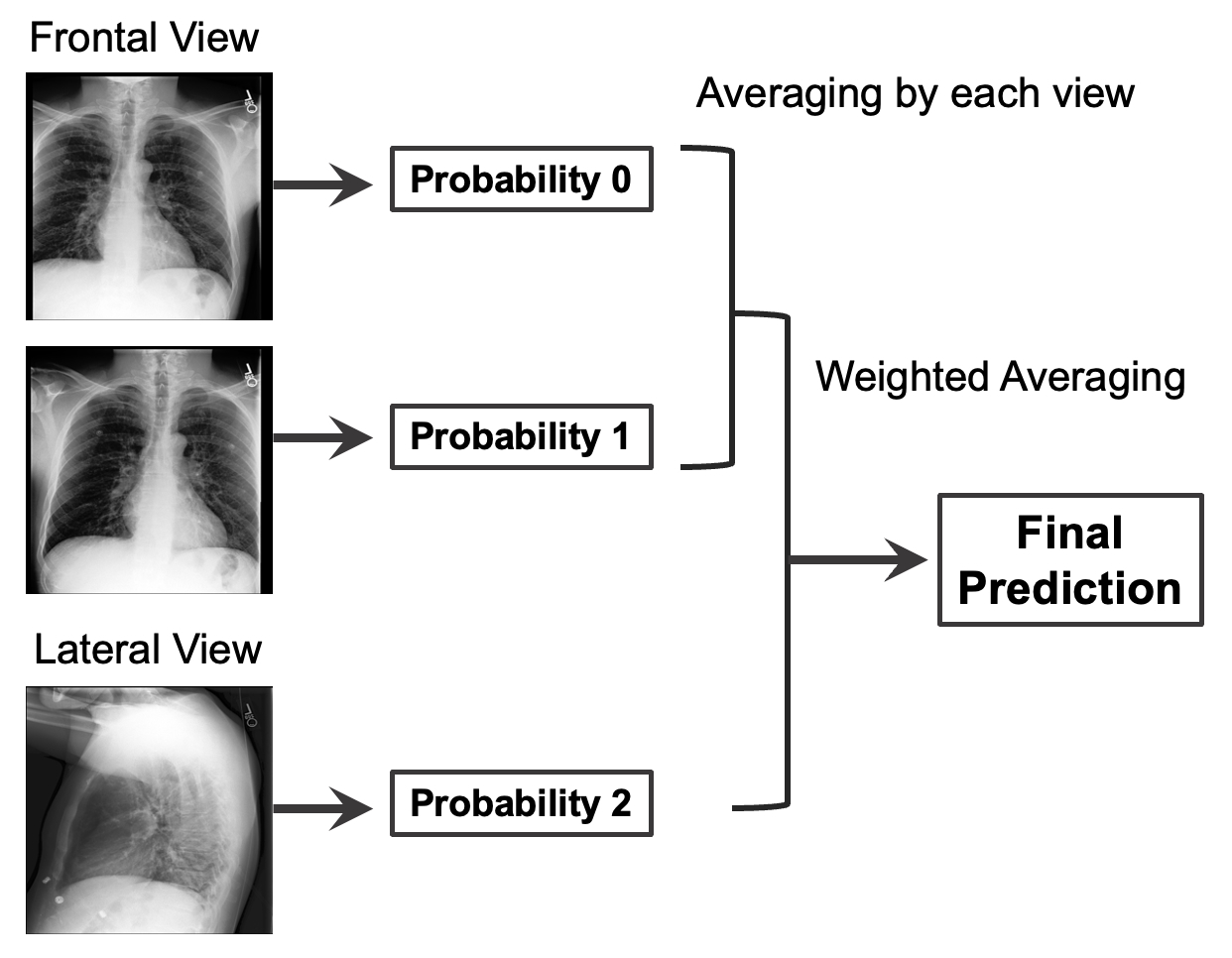}
    \caption{Multi-View Aggregation Process for Chest X-ray Prediction. 
    The diagram illustrates the method of combining predictions from multiple 
    chest X-ray views to produce a final prediction value. 
    Probabilities from frontal view(s) and lateral view(s) are processed 
    through view-specific averaging, followed by weighted averaging 
    across different views to generate the final prediction score.}
    \label{fig:cxrlt_fig1}
\end{figure}

This multi-view aggregation process is visually represented in Figure~\ref{fig:cxrlt_fig1}. The figure illustrates a two-stream approach where frontal and lateral chest X-ray images are processed separately through identical models.

Our aggregation strategy is implemented as follows:

\begin{enumerate}
    \item Generate predictions for each view independently.
    \item Calculate the mean prediction for frontal and lateral views separately:
    \begin{equation}
    P_f = \frac{1}{N_f} \sum_{i=1}^{N_f} P_{f,i}, \quad P_l = \frac{1}{N_l} \sum_{i=1}^{N_l} P_{l,i}
    \end{equation}
    where $P_f$ and $P_l$ are the mean predictions for frontal and lateral views respectively, $N_f$ and $N_l$ are the number of images for each view, and $P_{f,i}$ and $P_{l,i}$ are the predictions for individual images.
    \item Apply a weighted average to combine the mean predictions from frontal and lateral views:
    \begin{equation}
    P_{final} = \frac{w_f P_f + w_l P_l}{w_f + w_l}
    \end{equation}
    where $P_{final}$ is the final prediction, and $w_f$ and $w_l$ are the weights for frontal and lateral views respectively.
\end{enumerate}

\subsection{Data Preprocessing and Augmentation}
To ensure optimal model performance, we applied the following preprocessing steps to all images:

\begin{enumerate}
    \item Resizing: All images were resized to 384x384 or 512x512 pixels.
    \item Normalization: Pixel values were normalized to the range [0, 1] and then standardized using mean (0.485, 0.456, 0.406) and std (0.229, 0.224, 0.225) values from ImageNet.
\end{enumerate}

During training, we applied the following data augmentation techniques to improve model generalization:

\begin{itemize}
\item Random resized crop (scale: 0.85 to 1.0)
\item Random shift, scale, and rotation (probability: 0.5, rotation limit: ±15 degrees)
\item Random horizontal flipping (probability: 0.5)
\item Random cutout (max size: 5\% of image size, 5 holes, probability: 0.1)
\end{itemize}

\subsection{Training Process}
Our training process consisted of the following steps:

\begin{enumerate}
    \item Initialize models with weights pretrained on ImageNet.
    \item Pretrain on the NIH Chest X-ray dataset (MaxViT model only).
    \item Fine-tune on the MICCAI CXR-LT dataset using the asymmetric loss function.
\end{enumerate}
\textit{Note: The pretraining step in item 2 was applied only to the MaxViT model in our experiments.}

\subsubsection{Training Details}
\begin{itemize}
    \item Optimizer: AdamW with initial learning rate of 1e-4
    \item Learning rate schedule: Cosine annealing
    \item Batch size: 16
    \item Number of epochs: 5
\end{itemize}

\subsection{Evaluation Metric}
We used the mean Average Precision (mAP) as our primary evaluation metric, in accordance with the CXR-LT 2024 challenge. The mAP is particularly well-suited for multi-label classification tasks in imbalanced datasets, as it provides a single score that summarizes the precision-recall curve for each class.

The mAP is calculated as follows:

\begin{enumerate}
    \item For each class, compute the Average Precision (AP) by calculating the area under the precision-recall curve.
    \item Take the mean of the APs across all classes.
\end{enumerate}

This metric was used for model selection during both the pretraining phase and the final evaluation on the CXR-LT dataset. It provides a comprehensive assessment of the model's performance across all classes, giving equal importance to both common and rare conditions.

\section{Results}

\subsection{Model Architectures and Input Size}
We experimented with different input sizes, comparing 384 and 512. Across all model architectures, increasing the input size to 512 resulted in improved performance, suggesting that higher-resolution images provide more detailed information for the model to learn. Notably, MaxViT outperformed ConvNeXt V2 in most scenarios, demonstrating its superior capability in capturing complex features in chest X-ray images. Table \ref{tab:model-comparison} presents a detailed comparison of the performance metrics for different model architectures and input sizes.

\begin{table}[h]
\caption{Performance Comparison of Different Model Architectures and Input Sizes. Performance metrics for Subtask 1 and Subtask 2 are reported as mAP. Bold values indicate the best performance for each subtask.}
\centering
\renewcommand{\arraystretch}{1.2} % Increase row height
\begin{tabular}{p{3.5cm} p{1.5cm} p{3.5cm} p{2cm} p{2cm}}
\hline
\textbf{Model} & \textbf{Size} & \textbf{Pretrained} & \textbf{Subtask 1} & \textbf{Subtask 2} \\
\hline
ConvNeXtV2 Tiny & 384 & ImageNet-22K + 1K & 0.2549 & 0.3418 \\
MaxViT Tiny & 384 & ImageNet-1K & 0.2557 & 0.3418 \\
\hline
ConvNeXtV2 Tiny & 512 & ImageNet-22K + 1K & 0.2594 & 0.3464 \\
MaxViT Tiny & 512 & ImageNet-1K & \textbf{0.2621} & \textbf{0.3469} \\
\hline
\end{tabular}
% \caption{Comparison of Model Performance}
\label{tab:model-comparison}
\end{table}

\subsection{Effect of CXR Pretraining}
We evaluated the impact of using NIH CXR pretraining on MaxViT Tiny with an input size of 512. The results showed an improvement of approximately 0.001 to 0.002 in mAP for both subtasks when NIH-pretrained weights were used, as shown in Table \ref{tab:nih-cxr-comparison}. This improvement indicates that domain-specific pretraining is effective for enhancing model sensitivity to thoracic findings, particularly for those with subtle manifestations.

\begin{table}[h]
\caption{Performance Comparison of MaxViT Tiny (512) with and without NIH CXR Pretraining. Performance metrics for Subtask 1 and Subtask 2 are reported as mAP. Bold values indicate the best performance for each subtask.}
\centering
\renewcommand{\arraystretch}{1.2} % Increase row height
\begin{tabular}{p{3.5cm} p{2cm} p{2cm}}
\hline
\textbf{NIH Pretrained} & \textbf{Subtask 1} & \textbf{Subtask 2} \\
\hline
No & 0.2667 & 0.3524 \\
Yes & \textbf{0.2678} & \textbf{0.3545} \\
\hline
\end{tabular}
% \caption{Performance Comparison of MaxViT Tiny (512) with and without NIH CXR Pretraining}
\label{tab:nih-cxr-comparison}
\end{table}

\subsection{Effect of View-Based Prediction Aggregation}
Our view-based prediction aggregation strategy significantly improved performance for both subtasks, as demonstrated in Table \ref{tab:combined-results}. Aggregating predictions across views helps the model make more informed study-level decisions, especially when key findings are not clearly visible in a single view. The experiments confirmed that a weighted aggregation favoring the frontal view (e.g., 7:3 or 8:2) further boosts performance, aligning with the clinical observation that lateral views often provide less diagnostic information.

\begin{table}[h]
\caption{Performance Comparison of Different Prediction Aggregation Strategies. PP Ratio represents the weighting between frontal and lateral views. mAP values are reported for both subtasks across different models and datasets.}
\centering
\begin{tabular}{p{3cm} p{2.5cm} c @{\hspace{1cm}} p{1.5cm} @{\hspace{1cm}} c}
\toprule
Dataset & Model & Subtask & PP Ratio & mAP \\
\midrule
& & \multirow{2}{*}{1} & 5:5 & 0.2621 \\
& &  & 7:3 & 0.2667 \\
\cmidrule(lr){3-5}
& \multirow{-3}{*}{MaxViT Tiny} & \multirow{2}{*}{2} & 5:5 & 0.3469 \\
& &  & 7:3 & 0.3524 \\
\cmidrule(lr){2-5}
& & \multirow{2}{*}{1} & 7:3 & 0.2677 \\
& &  & 8:2 & 0.2680 \\
\cmidrule(lr){3-5}
\multirow{-7}{*}{Development} & \multirow{-3}{*}{Ensemble} & \multirow{2}{*}{2} & 7:3 & 0.3549 \\
& &  & 8:2 & 0.3560 \\
\midrule
\multirow{2}{*}{Test} & \multirow{2}{*}{Ensemble} & \multirow{2}{*}{2} & None & 0.4908 \\
& &  & 8:2 & 0.5065 \\
\bottomrule
\end{tabular}
\label{tab:combined-results}
\end{table}

\subsection{Test Dataset Performance}
On the test dataset, the ensemble of ConvNeXt V2 and NIH-pretrained MaxViT achieved competitive results, securing 4th place in Subtask 2 and 5th place in Subtask 1. As shown in Table \ref{tab:test-performance}, including the NIH-pretrained MaxViT in the ensemble led to a noticeable increase in mAP, confirming the value of combining domain-specific pretraining with an ensemble strategy for handling diverse and long-tailed chest X-ray findings.

\begin{table}[h]
\caption{Performance Comparison of Different Ensemble Combinations on the Test Dataset. mAP values are reported for both subtasks. Bold values indicate the best performance for each subtask.}
\centering
\begin{tabular}{ccc}
\hline
\textbf{Model Combination} & \textbf{Subtask 1} & \textbf{Subtask 2} \\
\hline
MaxViT + ConvNeXtV2 & 0.2722 & 0.5065 \\
MaxViT $\times$ 2 (w \& w/o NIH) + ConvNeXtV2 & \textbf{0.2728} & \textbf{0.5089} \\
\hline
\end{tabular}
\label{tab:test-performance}
\end{table}

\section{Discussion}

The results of our experiments demonstrate that ensembling state-of-the-art models, equipped with domain-specific pretraining, can effectively tackle the challenges posed by long-tailed CXR classification. The use of asymmetric loss proved essential in improving detection sensitivity for rare findings, addressing a common issue in medical image classification. Additionally, our view-based aggregation method emphasizes the importance of multi-view integration, a practice routinely employed by radiologists in clinical settings. However, this study did not explore alternative loss functions, suggesting a potential area for further research. Future work could involve investigating the impact of different loss functions and exploring advanced aggregation techniques to further enhance model performance.

\section{Conclusion}

This paper presents a novel approach to long-tailed CXR classification using an ensemble of ConvNeXt V2 and MaxViT models. By incorporating asymmetric loss and view-based aggregation, our method effectively improves detection accuracy, particularly for rare findings. Our results underscore the importance of model pretraining and multi-view integration in medical image classification tasks. Further exploration of different loss functions and aggregation strategies could provide additional gains in performance.

\begin{credits}
\subsubsection{\ackname} 
We acknowledge the MICCAI Challenge organizers and the community for providing the opportunity to participate in this task.

\subsubsection{\discintname}
There is no conflict of interest to disclose.

\end{credits}

% \bibliography{references}

\end{document}